%% file: main-InvSearch-CoRR.tex
\documentclass[letterpaper]{article}
\pdfoutput=1
\usepackage{aaai}
\usepackage{times}
\usepackage{helvet}
\usepackage{courier}
\newlength{\bibitemsep}
\newlength{\bibparskip}
\let\oldthebibliography\thebibliography
\renewcommand\thebibliography[1]{%
  \oldthebibliography{#1}%
  \setlength{\parskip}{\bibitemsep}%
  \setlength{\itemsep}{\bibparskip}%
}

\input{preface}

\input{macros}

\input{environments}

\title{Path planning with Inventory-driven Jump-Point-Search\thanks{Extended and revised version of {\protect \cite{AversaSardinaVassos:AIIDE15}}.}}

\author{
Davide Aversa \\
Department of Computer, Control, \\and Management Engineering\\
Sapienza University of Rome\\
Rome, Italy\\
\texttt{aversa@dis.uniroma1.it}
\And 
Sebastian Sardina\thanks{This work was conducted while a Visiting Professor at Sapienza Universit\`a di Roma, Rome, Italy.} \\
School of Computer Science\\ and Information Technology\\
RMIT University in Melbourne\\
Melbourne, Australia\\
\texttt{sebastian.sardina@rmit.edu.au }
\And 
Stavros Vassos \\
Department of Computer, Control, \\and Management Engineering\\
Sapienza University of Rome\\
Rome, Italy\\
\texttt{stavros@dis.uniroma1.it}
}
 
\begin{document}

\nocopyright
\maketitle

\begin{abstract}
In many navigational domains the traversability of cells is conditioned on the path taken. This is often the case in video\-games, in which a character may need to acquire a certain object (i.e., a key or a flying suit) to be able to traverse specific locations (e.g., doors or high walls).
In order for non-player characters to handle such scenarios we present \invJPS, an ``inventory-driven'' pathfinding approach based on the highly successful grid-based Jump-Point-Search (\JPS) algorithm.
We show, formally and experimentally, that the \invJPS preserves \JPS's optimality guarantees and its symmetry breaking advantages in inventory-based variants of game maps.
\end{abstract}

\input{intro}
\input{preliminaries}
\input{inventoryplanning}

\vspace{-.2cm}
\input{experiments}
%
\vspace*{-.2cm}
\input{related}
\input{conclusions}
%

{\small

\vspace{-0.2cm}

}

\newpage
\input{appx-proof}

\end{document}

%% file: preface.tex
\usepackage{enumitem}
\usepackage{times}
\usepackage{helvet}
\usepackage{courier}

\usepackage{latexsym}
\usepackage{amsmath}
\usepackage{amssymb}

\usepackage{multicol}
\usepackage{comment}
\usepackage{textcomp}
\usepackage{xspace}
\usepackage{subfigure}
\usepackage{soul}\setuldepth{x}
\usepackage{url}

\usepackage{tikz,pgf}
\usetikzlibrary{arrows,decorations,backgrounds,matrix,automata,shapes,shadows,trees,shapes,patterns,calc,fit,positioning}
\pgfdeclarelayer{background}
\pgfdeclarelayer{foreground}
\pgfsetlayers{background,main,foreground}
\usepackage{pgfplots}
\usetikzlibrary{pgfplots.groupplots}
\usepackage[amsmath,thmmarks]{ntheorem}
\theoremseparator{.}

\theorembodyfont{\itshape}
\newtheorem{theorem}{Theorem}
\newtheorem{proposition}{Proposition}
\newtheorem{lemma}{Lemma}

\theorembodyfont{\normalfont}
\theoremsymbol{\ensuremath{\blacksquare}}

\theoremsymbol{\ensuremath{\square}}

\theorembodyfont{\normalfont}
\theoremstyle{nonumberplain}
\theoremsymbol{\ensuremath{\square}}
\theoremheaderfont{\normalfont\itshape}
\newtheorem{proofsk}{Proof Sketch}

\newtheorem{proof}{Proof}

%% file: macros.tex
\newcommand{\mathname}[1]{\ensuremath{\text{\textit{#1}}}}
\newcommand{\propername}[1]{\text{\textsf{\small #1}}\xspace}

\newcommand{\astar}{\propername{A*}}
\newcommand{\JPS}{\propername{JPS}}
\newcommand{\HPA}{\propername{HPA*}}

\newcommand{\invAStar}{\propername{InvA*}}
\newcommand{\invJPS}{\propername{InvJPS}}

 \newcommand{\D}{\mathcal{D}}

\newcommand{\I}{\mathcal{I}}

\newcommand{\tup}[1]{\langle #1\rangle}            %
\newcommand{\tuple}[1]{\tup{#1}}            %

\newcommand{\myi}{\emph{(i)}\xspace}
\newcommand{\myii}{\emph{(ii)}\xspace}
\newcommand{\myiii}{\emph{(iii)}\xspace}
\newcommand{\myiv}{\emph{(iv)}\xspace}

\newcommand{\defterm}[1]{\ul{\textit{#1}}}	%

\newcommand{\obj}{\mathname{obj}}
\newcommand{\req}{\mathname{req}}
\newcommand{\adj}{\mathname{adj}}

%% file: environments.tex
\usepackage[amsmath,thmmarks]{ntheorem}
\theoremseparator{.}

\theorembodyfont{\itshape}

\theorembodyfont{\normalfont}
\theoremsymbol{\ensuremath{\blacksquare}}
\renewtheorem{definition}{Definition}

\theoremsymbol{\ensuremath{\square}}
\renewtheorem{example}{Example}

\theorembodyfont{\normalfont}
\theoremstyle{nonumberplain}
\theoremsymbol{\ensuremath{\square}}
\theoremheaderfont{\normalfont\itshape}
\renewtheorem{proofsk}{Proof (Sketch)}
\renewtheorem{proofapx}{Proof}
\renewtheorem{proof}{Proof}

\usepackage[textsize=scriptsize,textwidth=1.7cm]{todonotes}
\usepackage{cooltooltips}
\usepackage{etoolbox}
\makeatletter
\patchcmd{\@addmarginpar}{\ifodd\c@page}{\ifodd\c@page\@tempcnta\m@ne}{}{}
\makeatother
\reversemarginpar

%% file: intro.tex
\vspace*{-2ex}
\section{Introduction}\label{sec:intro}

Pathfinding is a fundamental problem that arises in many diverse scenarios ranging from robots in the real world, e.g., \cite{bruceveloso02randpath}, to videogame characters in virtual worlds, e.g., \cite{BoteaMulerSchaeffer:JCD04-HPA}, \cite{bjornsson2006improved}.
In all cases, an autonomous (physical or virtual) agent needs to reason over a map and find out how to reach a desired destination.

In videogames, in particular, it is often necessary to consider different capabilities for the characters that are performing pathfinding. For instance, some characters may be able to climb walls, fly, or swim, while others may not.
Still, in those cases, the capabilities of the agents are \emph{fixed}.
Consider instead  the scenario in which a non-player character (NPC) in a videogame is chasing the human player; the player goes through a corridor and locks the door behind him. With the corridor locked, the NPC has no other way to reach the player; nonetheless, there is a key, reachable by the NPC, that can unlock the door.
Clearly, the NPC can reach the player if it first gets to the location of the key, picks it up, and then \emph{goes back} to the corridor to open the door. %
Regular pathfinding approaches would fail to find a path (as the door is locked and there is no other way to reach the player). Common solutions in games involve hard-coding the behavior of the character to reach for a particular object (e.g., a key or suit), thus sacrificing flexibility at run-time as well as increasing the effort for developing the intended interaction.

The issue with the above example is that the agent's capabilities (in the broad sense of allowing them to traverse locations) are \emph{not} fixed at the outset, but \emph{depend on the path that is followed by the agent}.
Technically, whether a location is ``blocked'' or not, depends on whether the agent has visited a given location before and has \emph{acquired then certain items or capabilities}.
We shall refer to this variant of path planning as \emph{inventory-driven pathfinding}.

Of course, one possibility for handling such settings is to employ general task-planning methods, such as classical STRIPS/PDDL planning~\cite{GhallabNT:04-Planning}. By doing so, navigation, collecting objects, and using objects to open a blocked location, can be easily represented as actions with appropriate precondition and effects.
However, as we will show, this turns out to be an impractical approach to the task and is not suitable for applications like videogames, for which pathfinding requests ought to be resolved almost instantly.
We shall also point out that we do not aim at developing a complete new pathfinding approach for the inventory-driven task, but to rely on existing successful  techniques as much as possible---\emph{the simpler, the better}.
Our proposal is to build upon one of the most successful pathfinding algorithms, namely, Jump-Point-Search (\JPS)~\cite{HaraborGrastien:AAAI11-JPS,HaraborGrastien:SOCS12-JPS,HaraborGestien:ICAPS14-JPS}, and parsimoniously extend it to handle inventory-driven scenarios without losing its optimality properties nor, hopefully, its practical performance.
More concretely, we present \invJPS, an ``inventory-driven'' variant of \JPS that preserves the symmetry breaking advantages of \JPS  in the extended setting.
We evaluate the approach over synthetic and real videogame maps that are augmented with doors and keys.

%% file: preliminaries.tex
\section{Planning via Jump Points}\label{sec:preliminaries}

\input{fig-jps}

Jump-Point-Search (\JPS) is a recent technique introduced by Harabor and Grastien in  \shortcite{HaraborGrastien:AAAI11-JPS} and further elaborated in \shortcite{HaraborGrastien:SOCS12-JPS,HaraborGestien:ICAPS14-JPS}, that has proven extremely successful to navigate uniform cost grid-based maps (see~\cite{Masters:HON14} for initial efforts to generalize it to non-uniform grids).
\JPS is ultimately a speedup approach based on the well-known \astar \cite{HartNilssonRaphael:IEEE_TSSC68-ASTAR},
a heuristic best-first search algorithm to find a minimum-cost path through a graph.
Below, we shall assume that the reader is familiar with \astar and describe the working of \JPS to understand our contribution.
\vspace*{-2ex}
\paragraph{Overview}

\JPS operates by identifying every node that might end up being on the optimal path, i.e., the so-called \emph{jump-points}, and discards the rest.
Thus, instead of expanding all reachable nodes, \JPS jumps from a potential turning (jump) point to another turning (jump) point, expanding only those nodes that might require a change of direction. %

Following~\cite{Masters:HON14}, \JPS operates, technically, like \astar by working through the grid systematically, maintaining an open list, and selecting and opening nodes from that list using the same best-first evaluation function as A*; but \JPS is also a beam search in that it prunes as it goes.
When a node $x$ in the open list is expanded, \astar retrieves \emph{all} its unblocked adjacent nodes and adds them to the open list. 
Instead, \JPS retrieves only some adjacent nodes of $x$, constructs a \emph{vector of travel} from $x$ through each of them, and identifies the first so-called ``\emph{jump point}'' that occurs along that vector---a final successor of $x$---which is finally added to the open list. Informally, a jump point represents a location in which the traveling direction in the optimal path may change. 
Since intermediate nodes from node $x$ to each jump point are not explicitly handled or stored (in the closed list), once the goal is found and the path needs to be assembled, \JPS must ``fill in the gaps" between jump-points to generate a straight or diagonal path.

\newcommand{\neighbours}{\mathname{neighbours}}
\newcommand{\parent}{\mathname{parent}}

\vspace*{-3ex}
\paragraph{Pruning Mechanism}

Given a node $x$ reached via a parent node $p$, \JPS prunes from any node $n \in \neighbours(x)$ of $x$ such that (assuming corner-cutting is not allowed) either:%
\begin{enumerate}\addtolength{\itemsep}{-.8ex}
     \item there exists a path $\pi'$ from $p$ to $n$ that does not go through $x$ and is shorter than path $\pi = \tuple{p,x,n}$; or
     
     \item there exists a path $\pi'$ from $p$ to $n$ that does not go through $x$, has the same length as $\pi = \tuple{p,x,n}$, but $\pi'$ has a diagonal move earlier than $\pi$.
\end{enumerate}

For example, in Figure~\ref{fig:jps-straight} and \ref{fig:jps-diagonal}, pruned neighbours of $x$ are marked in gray; the remaining neighbours of $p$ are marked as white and are referred to as the \defterm{direct successors} of node $x$.
The \defterm{natural successors} of $x$ are the direct successors if one were to assume no obstacles whatsoever.
In the absence of obstacles, non-natural successors are pruned. But the presence of obstacles may preclude the pruned rules to discard some non-natural successor, which therefore end up in the set of direct successors. Such non-natural successors that could not be pruned form the set of \defterm{forced neighbours} of $x$. 
In Figure~\ref{fig:jps-forced}, the patterned cells $z_1$ and $z_2$ are forced neighbours of $x'$: they are not natural successors of $x'$, but they could not be pruned because the move $\tuple{p',z_1}$ cuts corner and is hence not legal.  
Importantly, only straight movements may produce (up to $4$) forced neighbours.

\vspace*{-3ex}
\paragraph{Jumping Procedure}
As explained by~\cite{HaraborGrastien:SOCS12-JPS}, \JPS applies to each actual successor--natural or forced neighbour---of the current node $x$ a simple ``jumping" procedure to replace each neighbour $n$ with an alternative successor $n'$ that is further away, the next ``jump point.''
Technically, a \defterm{jump point} is a node that contains a forced neighbour.
In Figure~\ref{fig:jps-forced}, node $x'$ is a jump point. Intuitively, the fastest way to reach $z_1$ and $z_2$ from $p'$ is via $x'$, and hence node $x'$ becomes a ``turning point.''

When moving in a straight manner, the only natural neighbour is explored recursively, in the corresponding straight direction, until either an obstacle, a jump point (i.e., a node having a forced neighbour; e.g., node $x'$ above), or the goal is encountered. In the first case, the path is deemed failed, all nodes in it are ignored, and nothing is generated. If, however, a node $n'$ having one or more forced neighbours---jump points---or being the actual goal is reached, then $n'$ is generated as a next successor of $x$ and is added to the open list; effectively ``jumping'' from $x$ to $n'$ without adding any of the intermediate nodes into the open list.
So, in Figure~\ref{fig:jps-forced}, the final successor of node $x$ is jump point node $x'$, which has nodes $z_1$ and $z_2$ as a forced neighbours. 

When moving diagonally, as in Figure~\ref{fig:jps-diagonal}, node $x$ has three natural neighbours (white cells in $x$ surroundings): two straight (north and east) and one diagonal (east north). \JPS then recourses over the diagonal neighbour only if both straight neighbours produce failed paths. When there is a non-failed straight jump, then the node in the diagonal path is also considered an (indirect) jump point---a potential turning point---and is added to the open list (node $y$).

By jumping, \JPS reduces memory consumption and the number of operations required. \JPS was shown to improve on the fastest pathfinding approaches (including \HPA) by several factors in some cases~\cite{HaraborGrastien:AAAI11-JPS}.
\JPS operations are all done online, with no pre-processing or memory overhead, and 
moreover, it is provably optimal.%

%% file: fig-jps.tex
\pgfdeclarelayer{background}
\pgfdeclarelayer{foreground}
\pgfsetlayers{background,main,foreground}

\newcommand{\obstacle}[2]{
    \draw[draw=none,fill=black] (#1-1,#2-1) rectangle (#1,#2);
}

\newcommand{\akey}[3]{
	\gridlabel{#1}{#2}{\includegraphics[scale=.14,angle=-45]{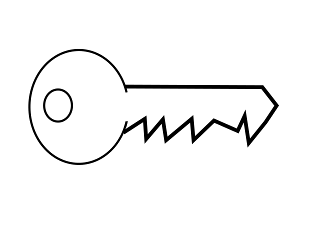}}{#3}
}

\newcommand{\door}[3]{
	\gridlabel{#1}{#2}{\includegraphics[width=1cm,height=1cm]{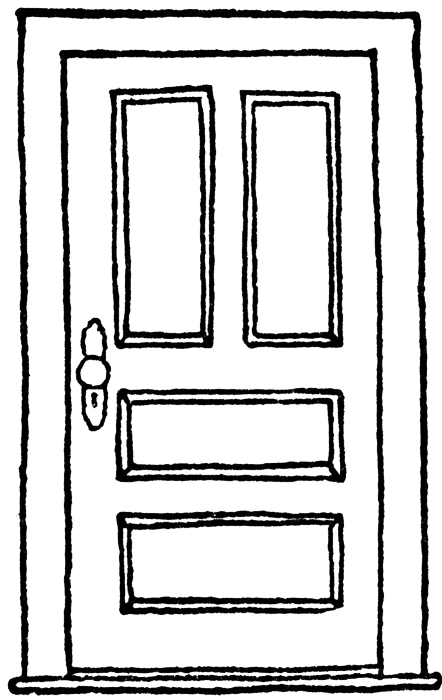}}{#3}
}

\newcommand{\pruned}[2]{
 \begin{pgfonlayer}{background}   
    \draw[draw=none,fill=black!20] (#1-1,#2-1) rectangle (#1,#2);
 \end{pgfonlayer}
}

\newcommand{\forced}[2]{
 \begin{pgfonlayer}{background}   
    \draw[draw=none,pattern=north east lines] (#1-1,#2-1) rectangle (#1,#2);
 \end{pgfonlayer}
}
\newcommand{\gridlabel}[4]{
    \node[draw=none,fill=none] at (#1-.5,#2-.5) (#4) {#3};
}
\newcommand{\gridarrow}[5]{
    \draw[draw=black,thick,#5] (#1-.3,#2-.5) -- (#3-.5,#4-.5);
}

\begin{figure*}[t]
\begin{center}
\subfigure[Straight movement from $x$ towards jump-point $x'$ (see (c)).]{
\label{fig:jps-straight}
\resizebox{.45\columnwidth}{!}{
  \begin{tikzpicture}
    \draw[black,line width=1pt] (0,0) grid[step=1] (7,5);
    \draw[black,line width=2pt] (0,0) grid[step=3] (3,3);

    \obstacle{5}{3}
    
    \gridlabel{1}{2}{$p$}{p}
    \gridlabel{2}{2}{$x$}{x}
    \gridlabel{6}{2}{$x'$}{y}

    \draw[very thick,fill=black,->] (p) -- (x);
    \draw[very thick,fill=black,dotted,->] (x) -- (y);

	\pruned{1}{1}
	\pruned{1}{2}
	\pruned{1}{3}
	\pruned{2}{1}
	\pruned{2}{3}
	\pruned{3}{1}
	\pruned{3}{3}
     
  \end{tikzpicture}
}
}~~~
\subfigure[Diagonal movement from $x$. Node $y$ is a jump-point.]{
\label{fig:jps-diagonal}
\resizebox{.45\columnwidth}{!}{
  \begin{tikzpicture}
    \draw[black,line width=1pt] (0,0) grid[step=1] (7,5);
    \draw[black,line width=2pt] (0,0) grid[step=3] (3,3);
    \obstacle{6}{3}
    \obstacle{6}{2}
    
    \gridlabel{1}{1}{$p$}{p}
    \gridlabel{2}{2}{$x$}{x}
    \gridlabel{4}{4}{$y$}{y}
    \gridlabel{7}{4}{$x'$}{z}
    
    \draw[very thick,fill=black,->] (p) -- (x);
    \draw[very thick,fill=black,->] (x) -- (y);
    \draw[very thick,fill=black,->] (y) -- (z);

    \gridlabel{2}{5}{}{x1}
    \draw[thick,fill=black,->,dashed] (x) -- (x1);
    \gridlabel{5}{2}{}{x2}
    \draw[thick,fill=black,->,dashed] (x) -- (x2);

    \gridlabel{3}{3}{}{xn}
    \gridlabel{3}{5}{}{x3}
    \draw[thick,fill=black,->,dashed] (xn) -- (x3);
    \gridlabel{5}{3}{}{x4}
    \draw[thick,fill=black,->,dashed] (xn) -- (x4);

    \gridlabel{4}{5}{}{x5}
    \draw[thick,fill=black,->,dashed] (y) -- (x5);

	\pruned{1}{1}
	\pruned{1}{2}
	\pruned{1}{3}
	\pruned{2}{1}
	\pruned{3}{1}

  \end{tikzpicture}
}
}~~~
\subfigure[Two forced neighbours $z_1$ and $z_2$ of jump-point $x'$.]{
\label{fig:jps-forced}
\resizebox{.45\columnwidth}{!}{
  \begin{tikzpicture}
    \draw[black,line width=1pt] (0,0) grid[step=1] (7,5);
    \obstacle{5}{3}
    
    \gridlabel{1}{2}{$p$}{p}
    \gridlabel{2}{2}{$x$}{x}
    \gridlabel{5}{2}{$p'$}{x1}
    \gridlabel{6}{2}{$x'$}{y}
    \gridlabel{6}{3}{$z_1$}{z1}
    \gridlabel{7}{3}{$z_2$}{z2}
    
    \draw[very thick,fill=black,->] (p) -- (x);
    \draw[very thick,fill=black,->,dotted] (x) -- (x1);
    \draw[very thick,fill=black,->] (x1) -- (y);

	\pruned{5}{1}
	\pruned{5}{2}
	\pruned{5}{1}
	\pruned{6}{1}
 	\pruned{7}{1}
	\forced{6}{3}
	\forced{7}{3}
    \draw[black,line width=2pt] (4,0) rectangle (7,3);
     
  \end{tikzpicture}
}
}~~~
\subfigure[A scenario where one key can open two doors.]{
\label{fig:key-case}
\resizebox{.45\columnwidth}{!}{
  \begin{tikzpicture}
    \draw[black,line width=1pt] (0,0) grid[step=1] (7,5);
    \obstacle{5}{4}
    \obstacle{6}{4}
    \obstacle{7}{4}
    \door{5}{5}{d1}
    \gridlabel{6}{5}{$B$}{B}

    \obstacle{2}{3}
    \obstacle{2}{4}
    \obstacle{2}{5}
    \door{1}{3}{d2}
    
    \gridlabel{2}{2}{$p$}{p}
    \gridlabel{3}{2}{$x$}{x}
    \gridlabel{1}{4}{$A$}{A}
    \gridlabel{1}{2}{$y$}{y}

    \akey{6}{2}{k}

    \draw[very thick,fill=black,->] (p) -- (x);
    \draw[very thick,fill=black,->] (x) -- (k);

  \end{tikzpicture}
}
}
\end{center}
\vspace*{-3.3ex}
\caption{Four scenarios describing \JPS's straight and diagonal jumping mechanism and a key-door situation.}
\label{fig:jps-i}
\vspace*{-2ex}
\end{figure*}
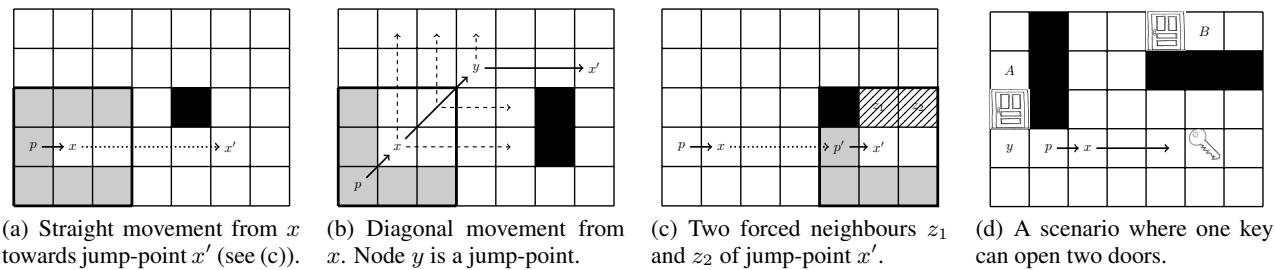

%% file: inventoryplanning.tex
\section{Inventory-driven Path Planning}\label{sec:inv-path-planning}

Here, we are interested in path planning \emph{in the context of agents who carry objects}---an ``inventory''---\emph{that can influence the navigation process}. In such settings, reaching the destination may depend on whether the agent has acquired a certain object, such as a key to open a door, or a swimming suit or boat to pass across water, as is typical in many videogame worlds. We will use ``items'' to refer to objects or capabilities that can be acquired by the agent, and can be used to traverse or open some blocked nodes in the map. %

\medskip
We define \emph{inventory-driven pathfinding} as follows. Given:
\begin{itemize}\addtolength{\itemsep}{-.8ex}
\item a grid-based map $M$ as a set of nodes $\{m_{11},m_{12},\ldots\}$;

\item a set $O \subseteq M$ of nodes that are blocked (and cannot be traversed by the agent);

\item a function $\adj: M \mapsto 2^M$ denoting the adjacency relation among nodes ($\adj(x)$ denotes the set of nodes that adjacent to node $x$);

\item a set of items $\I$ (e.g., objects or capabilities) that may be scattered in the map (and that the agent is able to acquire when co-located);

\item a function $\obj: M \mapsto 2^\I$ stating the items present in each location ($\obj(x) = \emptyset$ denotes no items at node $x$);

\item a function $\req: M \mapsto 2^\I$ stating which items are required to traverse a node; and

\item a start node $S$ and destination node $G$,
\end{itemize}
find an \emph{optimal (i.e., shortest) path} $\sigma = x_1,x_2,\ldots,n_n$, with $n \geq 1$, such that \myi $x_1 = S$; \myii $x_n=G$; \myiii $x_{i+1} \in \adj(x_i)$ and $x_i \not\in O$, for every $i \in \{1,\ldots,n-1\}$; and \myiv $\bigcup_{j < i} \obj(x_j) \subseteq \req(x_i)$, for every $i \in \{1,\ldots,n\}$
That is, we are interested in finding the shortest path from start to destination under the constraint that some locations along the path may require the agent to have previously visited other special nodes (fourth constraint above).

We note that this is a \emph{one-shot complete-knowledge} task, in the sense that the agent has all the relevant information at the outset---the map nodes and their connectivity, the blocked nodes, the items at nodes, and the items that open blocked nodes---and will deliberate \emph{offline} in order to find the shortest path that brings her to destination.
Different strategies for re-planning may be employed to account for a dynamic map or other variants for incremental planning, e.g.,\cite{Koenig:2005uq,koenig2004lpa}, but these are out of the scope of this work.

While there are various straightforward solutions for dealing with pathfinding in videogames in the context of characters with different capabilities, e.g., characters that can fly or swim, and characters with different sizes, there is no documented approach for dealing with this type of inventory-driven pathfinding where objects and capabilities can be acquired in the course of executing the path.
Similarly, in the academic literature for pathfinding search, there is no approach, to the best of our knowledge, that handles this particular variant of pathfinding. Of course, inventory-driven pathfinding can be formalized as a planning problem, e.g., by appealing to classical STRIPS planning~\cite{GhallabNT:04-Planning}, and using actions with appropriate precondition and effects in order to represent navigation, collecting objects, and using objects to open a blocked location. However, as we will show, being a domain-independent approach, this is an impractical approach for this task and is not suitable for applications like videogames, for which pathfinding requests ought to be resolved almost instantly.

Finally, we note that it is not possible, in general, to know in advance if an inventory item in the map will be needed or not in order to traverse the optimal path. This is indeed something to be discovered as part of the planning process and, as such, no item may in general be ignored at the outset.%

\subsection{Inventory-driven Jump-Point-Search}

As we are motivated by videogame worlds, we looked into academic techniques that are most influential in the practical videogame setting. \JPS is an award-winning algorithm and has attracted much attention within the game community,  %
and as all other pathfinding approaches is not able to cope with such inventory-driven scenarios: it will either yield the best path to the goal that does not resort to acquiring extra capabilities or output no solution. In this work then we show that \JPS can be further elaborated in a principled way to accommodate inventory-driven path planning.
The new algorithm, which we call \emph{inventory-driven \JPS} (\invJPS) is obtained by modifying \JPS in three simple ways.

The \textbf{first modification}, as one would do with any search approach, involves extending the state representation to account not just for the location of the agent, but also the current inventory. So, for a map $M$ a state is a pair $\tuple{x,I}$ where $x$ is a node in $M$ and $I$ is a subset of elements from set $\I$ of all possible items. As in this work we do not consider the cost for obtaining or carrying an item, during search when an agent is at a location that has items, these are placed all in their inventory instantly. The items that an agent carries allow them to traverse also nodes that are marked as blocked but are labeled by $\req$ with a set of items such that the intersection with the inventory $I$ is non-empty.

The \textbf{second modification} to \JPS involves treating any node containing some capability or object as an ``intermediate'' goal.
In other words, during the (recursive) jumping process, when an \emph{intermediate} node $x$ is
generated such that $\obj(x) \not= \emptyset$, then the process is deemed complete and $x$
is considered a jump point (and thus added to the open list).
We call such jump point nodes, \emph{inventory jump points}. Note that when an item $i$ is already contained in the state then acquiring another instance of the same item $i$ (by visiting a node where the second instance lies) does not change the state representation in the search process and does not generate an inventory jump point. In the case though that two items $i,i'$, such that $i\neq i$ happen to unlock the same nodes in the map, acquiring the essentially ``duplicate'' item $i'$ is handled in the generic way by the algorithm.
Finally, observe that, contrary to what one would expect, we do not require any change to the
pruning mechanism and hence we took a ``lazy'' approach to inventory ``finding.'' %

In classical \JPS, when a new (jump point) node is retrieved from the open list for expansion, \emph{all} directions towards natural and forced neighbours of the node in question are considered for further ``jumps.''
The natural successors represent the same direction the agent was traversing when a jump point was found, whereas the forced neighbours represent the turning directions that the agent may need to consider (see~\cite[Lemma 1]{HaraborGrastien:AAAI11-JPS}). This implies, for example, that it is not necessary to consider the parent of the jump node, as this would involve undoing the path traveled so far. %
However, in the context of inventory-based path planning, if the node being expanded is an inventory jump point, then the agent ought to consider \emph{all possible directions}, including that undoing the path traversed so far! The fact is that, with the new inventory acquired, nodes that looked ``blocked'' before may have now become traversable.

So, the \textbf{third modification} included in \invJPS is to treat inventory jump point nodes as the starting node, thus applying the jumping process towards all possible directions.
This is depicted in Figure~\ref{fig:key-case} (assume the key opens all doors). When jumping east from $x$, the node with the key becomes a jump point. From there, the agent must consider jumping towards \emph{all} directions, not just east. In fact, the agent should consider returning back to node $x$ as a new jump point, now though holding the key. Later, the agent will possibly generate node $y$ as a jump point too, because the node north to it is now open to the agent and will allow her to visit area $A$. A similar argument can be given for the north door and area $B$ (using the north-west direction).

Let us refer with \invJPS to the algorithm obtained from implementing the above three modifications to \JPS. %
The following result states that the optimality of \JPS is preserved.

\begin{theorem}\label{thm:optimality}
\invJPS always returns an optimal solution for any inventory-driven path planning problem.
If there is a path from start to destination, \invJPS returns a solution.

\begin{proofsk}
This can be proved by following the same reasoning as the proof of~\cite[Theorem 1]{HaraborGrastien:AAAI11-JPS}, but accounting for inventory locations that can allow arbitrary changes in direction of travel. See Appendix.
\end{proofsk}
\end{theorem}

\input{performance-analysis}

%% file: performance-analysis.tex
In fact, it is easy to see that when applied to non-inventory path planning problems, \invJPS reduces to regular \JPS.

\begin{theorem}\label{thm:same_expanded}
If \invJPS is applied to a path planning problem where $\obj(x)=\emptyset$, for all 
$x\in M$, then $\invJPS$ generates the same search space than \JPS, including the same 
list and order of nodes in the open list and intermediate expansions. 

\begin{proof}
As there are no nodes with objects in the map $M$ then: 
\emph{(i)} all search states are of the form $\tuple{x,\emptyset}$, hence 
the cost of navigating through a cell remains unchanged; and \emph{(ii)} no 
inventory jump point nodes are ever generated.
\end{proof}
\end{theorem}

\input{fig-naive-keys}

\subsection{Analysis of performance challenges for \invJPS}
\label{sec:badscenario}

For simplicity, we will refer to inventory nodes as nodes with \emph{keys} that open special (locked) \emph{door} nodes. A central observation is that, when \invJPS reaches a key, it considers going ``backwards,'' thus re-exploring previously explored areas in the hope of finding a now open door. 
Informally, the worst case scenario for \invJPS arises when the whole map is covered with keys such that each node contains a different key and the goal is unreachable, in which case the algorithm has to search the whole map multiple times, once for each possible combination of keys. 
However, this kind of scenario is unlikely to be seen in real games. Instead, we investigate how the number of keys in the map, their placements, and whether they are necessary for the optimal path, play an important role in how much ``re-exploration'' is performed.

Consider a map with only one key $k$ that is located at node $x_k$, and start and destination nodes $x_s,x_d$. Suppose first that there are no doors in the map, and so the key is not necessary for finding a path that reaches the destination. In such case, \invJPS faces a performance drawback scenario in the sense that one could simply apply standard \JPS. Motivated by this, we report below on experiments with redundant keys to evaluate the (sometimes) unnecessary overhead of \invJPS over \JPS. Note however that this is somehow an unfair comparison, because the whole point is that one does not know at the outset whether the keys are necessary or not.

Now, suppose further that the search reaches node $x_k$ (where the key is) and so generates the search state $\tuple{x_k,\{k\}}$. As all directions from there will be explored, the search could eventually be exploring concurrently search states of the form $\tuple{y,\{\}}$ and $\tuple{y,\{k\}}$, that is, the same location node $y$ but with different inventories (and different $g$ cost). However, the key is actually not needed. In this case, the \emph{earliest} that $x_k$ is encountered (e.g., if it is closest to the starting location), the \emph{more overhead} it will impose.

Now consider a map with only one key $k$ as before, but in this case there is also a 
locked door node $y$ such that all paths to the destination \emph{necessarily} pass through 
$y$. In this case the fact that $\tuple{x_k,\{k\}}$ is added to the open list is 
actually helpful, and it is more beneficial to happen early in the search process. 
As before, there will be nodes that are checked twice in different states of the form
$\tuple{y,\{\}}$ and $\tuple{y,\{k\}}$, even though the first is not necessary as there 
is no way to reach the destination without the key. Observe though that, unlike the 
cases where the key was \emph{not} needed, here the \emph{latest} that $x_k$ is encountered (e.g., if it is farthest from from the 
starting location), the \emph{more overhead} it will impose. For instance, consider 
the case in which the search needs to explore \emph{all the search space} 
before hitting the node with the key; then the search will essentially start over 
from $\tuple{x_k,\{k\}}$ to reach the destination going over nodes that have been 
explored in states with an empty inventory.

Finally, note that this scenario we described for the case that key $k$ is needed, 
is in fact very similar to the case in which a regular pathfinding search fails to find a 
path after exploring the whole search space. This suggests that we can test for 
``near-worst case scenarios'' when keys are needed, by means of checking \invJPS on 
\emph{unreachable} paths over maps with keys. This is clearly an upperbound for the expected 
overhead but also gives some qualitative insight.

%% file: fig-naive-keys.tex
\begin{figure*}[!th]
\begin{center}
\subfigure[Runtime of algorithms with keys randomly placed on the map.] {
\label{fig:exp1}
\resizebox{.23\textwidth}{!}{
\begin{tikzpicture}[-]
\begin{axis}[
	-,
	xmin = 0,
	xmax = 100,
	xlabel=Number of keys placed on the map,
	ylabel=Runtime in seconds,
	scaled y ticks=false,
	ymin=0,
	yticklabel style={
				/pgf/number format/fixed,
				/pgf/number format/precision=2,
	},
	y label style={at={(0.02,0.5)}},
	legend columns=-1,
	legend pos=north west,
	]
	\addplot [red, mark=square]
		table [y index=1] {data/EXP1-A.dat};
	\addplot [black, dashed, mark=x]
		table [y index=3] {data/EXP1-A.dat};
	\legend{\invJPS, \JPS}
\end{axis}
\end{tikzpicture}
}
}~~
\subfigure[Runtime of algorithms with keys placed on the path.] {
\label{fig:exp2a}
\resizebox{.23\textwidth}{!}{
\begin{tikzpicture}[-]
\begin{axis}[
	-,
	xmin = 0,
	xmax = 10,
	xlabel=Number of keys placed on the path,
	ylabel=Runtime in seconds,
	scaled y ticks=false,
	ymin=0,
	yticklabel style={
				/pgf/number format/fixed,
				/pgf/number format/precision=2,
	},
	y label style={at={(0.02,0.5)}},
	legend columns=2,
	legend pos=north west,
	]
	\addplot [red, mark=square]
		table [y index=1] {data/JPS-AVG-0-10.dat};
	\addplot [blue, mark=o]
		table [y index=1] {data/JPS-BEG-0-10.dat};
	\addplot [green, mark=x]
		table [y index=1] {data/JPS-END-0-10.dat};
	\addplot [black, mark=x, dashed]
		table [y index=5] {data/JPS-AVG-0-10.dat};
	\legend{\invJPS(MID), \invJPS(BEG), \invJPS(END), \JPS}
\end{axis}
\end{tikzpicture}
}
}~~
\subfigure[Speedup for the cases of (b) compared to \invAStar (log scale).] {
\label{fig:exp2b}
\resizebox{.23\textwidth}{!}{
\begin{tikzpicture}[-]
\begin{axis}[
	-,
	xmin = 1,
	xmax = 10,
	ymax = 100,
	xlabel=Number of keys placed on the path,
	ylabel=Speedup factor,
	ymode=log,
	scaled y ticks=false,
	ymin=0,
	yticklabel style={
				/pgf/number format/fixed,
				/pgf/number format/precision=2,
	},
	y label style={at={(0.02,0.5)}},
	legend columns=2,
	legend pos=north west,
	]
	\addplot [red, mark=square]
		table [y index=1] {data/EXP3SPEEDUP.dat};
	\addplot [blue, mark=x]
		table [y index=2] {data/EXP3SPEEDUPb.dat};
	\addplot [green, mark=o]
		table [y index=3] {data/EXP3SPEEDUP.dat};
	\legend{\invJPS(MID), \invJPS(BEG), \invJPS(END)}
\end{axis}
\end{tikzpicture}
}
}~~
\subfigure[Runtime of \invJPS with keys at the beginning of the path.] {
\label{fig:exp3}
\resizebox{.23\textwidth}{!}{
\begin{tikzpicture}[-]
\begin{axis}[
	-,
	xmin = 50,
	xmax = 355,
	xlabel=Path length,
	ylabel=Runtime in seconds,
	scaled y ticks=false,
	ymin=0,
	yticklabel style={
				/pgf/number format/fixed,
				/pgf/number format/precision=2,
	},
	y label style={at={(0.02,0.5)}},
	legend columns=2,
	legend pos=north west,
	]
	\addplot [red, mark=o]
		table [y index=11] {data/EXP4-JPS.dat};
	\addplot [red, mark=square]
		table [y index=2] {data/EXP4-JPS.dat};
	\addplot [blue, mark=o]
		table [y index=4] {data/EXP4-JPS.dat};
	\addplot [green, mark=x]
		table [y index=8] {data/EXP4-JPS.dat};
	\legend{0 keys, 2 keys, 4 keys, 8 keys}
\end{axis}
\end{tikzpicture}
}
}
\end{center}
\vspace*{-2.8ex}
\caption{Experimental results for the case that keys are not necessary over benchmark game maps.}
\label{fig:naive-experiments}
\vspace*{-2ex}
\end{figure*}
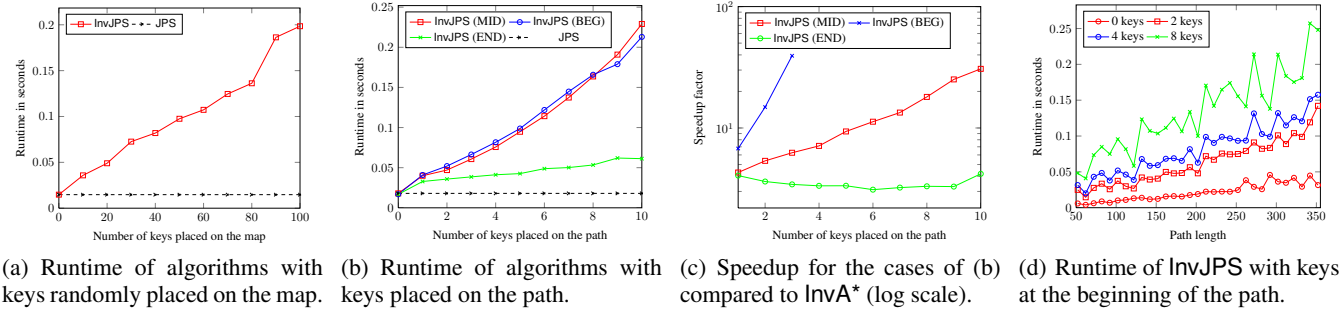

%% file: experiments.tex
\section{Experimental Analysis}\label{sec:experiments}

We report here on a series of experiments that demonstrate the performance of \invJPS applied to videogame maps from the \textsf{\small Moving AI} benchmark\footnote{\url{http://movingai.com/benchmarks/}}~\cite{sturtevant2012benchmarks} as well as synthetic maps we built.
\hspace{-.1cm}We use a Python implementation\footnote{Framework based on P4-simulator \url{http://tinyurl.com/p4-sim}; \cite{Masters:HON14}.} of the algorithms and run the experiments on an Intel i7 3.2GHz machine (on a single core) with 8GB of RAM. The runtime of the algorithms reported are often higher than the one that can be achieved by C++ implementations that would typically be employed in videogames.
Nonetheless, the particular implementation platform does not alter the core findings and conclusions and, of course, one would always resort to the most efficient platform at deployment  time.

In the first three experiments, we used the benchmark consisting of maps from ``Baldurs Gate II'' that are scaled to $512\times 512$ nodes. From each of the $75$ maps we took $50$ pathfinding instances that are known to be realizable and have a length between $150$ and $250$ steps.

\paragraph{Experiment 1: Random placement of \emph{unnecessary} keys over \emph{real} game maps; analysis per number of keys.}
First, we look into the case where regular maps are extended to include keys but not doors, so that the keys are \emph{not necessary} (and every key is different one from the another), and we compare to running regular \JPS. This experiment aims to evaluate the ``theoretical" overhead of \invJPS over \JPS. Effectively, the keys become ``noise'' for the planner, as discussed in the previous section.%

To extend the standard key-free maps, we added $0-100$ distinct keys in a random way in each map and run \invJPS and \JPS on those instances.
The runtime of the algorithms is reported in Figure~\ref{fig:exp1}. As expected, there is an overhead for considering keys, even when not needed, which scales well (almost linearly) over large numbers of keys. In particular, the overhead of running \invJPS over for the case of having $100$ keys on the map is equivalent to running approximately $10$ regular pathfinding queries with \JPS.
This is because the runtime is affected only by the keys which are actually taken into account during search. As we spread the keys uniformly on the map, not every node containing a key will be explored by the algorithm. 
$100$ keys is probably a large number for such maps, but it ensures that at least a few of those keys will appear in the random paths we try in this experiment. 

Another way to interpret the results is that what we see is in fact the start of an exponential overhead with respect to the small number of keys that are actually encountered during search in these random scenarios. The next experiment intends to explore this further by placing the keys in a positions that are relative to the shortest path solution.

\vspace*{-2.5ex}
\paragraph{Experiment 2: Placement of \emph{unnecessary} keys on the path over \emph{real} game maps; analysis per number of keys.}

Here, we consider again unnecessary keys but, following the discussion in the previous section, we look into the case that they are selectively inserted in three areas: \myi at the beginning of the path (BEG); \myii at the end of the path (END); and \myiii distributed evenly on the shortest path (MID). We assume that one tile can be occupied by at most one key, therefore, keys are scattered appropriately in a small area around the starting tile (BEG), the destination (END) or around the shortest path.
We tested \invJPS, plain \JPS, as well as  \invAStar, a straightforward inventory-driven variant of standard \astar.

As argued, the overhead is expected to increase when more keys are reached during search.
The runtime of \invJPS compared with that of \JPS (for reference) is reported on Figure~\ref{fig:exp2a}, while the speedup of \invJPS compared to \invAStar, in terms of how much faster is the runtime of \invJPS, is reported on Figure~\ref{fig:exp2b}. Note that every configuration (BEG, END and MID) is compared with \invAStar in the same configuration, e.g., \invJPS(BEG) is compared to \invAStar(BEG).

The results validate the expected effect of the placement of unnecessary keys on the runtime performance, with the (BEG) distribution being the worst configuration and the (END) distribution the best.
The exponential speedup we observe in the BEG scenario is due to the known speed-up of JPS over A*, but applied over all those conditioned cases of the search space for every combination of keys that is encountered. As the number of these combinations is exponential to the number of keys, the speedup of \invJPS then is also exponential. In fact, when we explore a smaller number of conditioned cases such as in the END scenario the exponential speedup is less evident (note that Figure~\ref{fig:exp2b} is in logarithmic scale).
These results validate that the benefits of \JPS are carried over in the inventory-driven setting. In particular, in the case of just $4$ keys close to the start location, \invJPS is more than $300$ times faster than \invAStar ($\invAStar$ is not able to cope with more than $4$ keys here).

\vspace*{-2.5ex}
\paragraph{Experiment 3: Placement of \emph{unnecessary} keys on the path over \emph{real} game maps; analysis per path length.}
In this experiment we test \invJPS over the most challenging scenario from the previous analysis, i.e., the BEG scenario, with respect to the path length.
We consider $100$ realizable paths with no length restriction per map. We then compute the optimal path and construct inventory-driven instances where keys are placed in the beginning of the path.
Figure~\ref{fig:exp3} shows the runtime of \invJPS wrt plan length for the cases in which $0$, $2$, $4$, and $8$ keys are added in the beginning of the path.
When no keys are added, \invJPS reduces to regular \JPS (Theorem~\ref{thm:same_expanded}). This can be used as reference to see the impact of the number of keys for different path lengths, compared to non-inventory paths of the same length that are solved with \JPS.
Observe that there is an approximately linear increase of the runtime of \invJPS as the path length increases.
While, as expected, more keys induce more effort to \invJPS, it is still able to solve problems with long paths, which would be impractical due to memory and time constraint for any \astar version doing complete node expansions.

\vspace*{-2.5ex}
\paragraph{Experiment 4: Incremental scenario of \emph{necessary} keys over \emph{synthetic} maps; analysis per number of keys.}
Here, we focus on the case where keys are actually needed in order to reach destination. As there is no benchmark maps for this setting, we built synthetic $512\times 512$ maps of two types starting from completely empty maps. Then, for each number of necessary keys ranging from $0$ to $10$, we report the average runtime of \invJPS over $200$ pathfinding instances.

In the first type of maps, we build artificial rooms that separate the empty map from left to right by putting vertical walls and a door between adjacent rooms. We impose a ``sequential'' traversal of the rooms in the sense that door $d_i$ cannot be reached without crossing door $d_{i-1}$,  and key for door $d_i$ can always be found in the map region before $d_i$. The destination is behind the last door.
\invJPS's performance is shown in Figure~\ref{fig:exp5a}.
For the second type, we impose a ``detour'' behavior.
For $n$ keys, we start from the destination, build a ``room'' around it, and then create a random entry point with a door $d_n$. We then chose a random point in the map, place key $k_n$ for door $d_n$, and repeat the process now with key $k_n$ (until all keys and doors are configured).
Unlike the first case, here doors are all reachable by the agent from the starting state, but solutions require the agent to go back and forth along all the map as the key for door $d_i$ is closed in the room blocked by door $d_{i-1}$ and so on. \invJPS's performance is shown in Figure~\ref{fig:exp5a}.

As expected, in the ``detour'' scenario the average path length ($450$-$500$ steps) is larger than in the ``sequential'' scenario as well as the other experiments, due to the multiple detours the agent has to perform. As the placement of rooms is done randomly and we only place a small number of rooms, the whole area of the map is not necessarily covered though, and this is why the average path length is not larger than the map dimensions.
Observe also that the runtime of \invJPS for $10$ keys in the ``sequential'' scenario is comparable to the runtime of \invJPS in Experiment 2 with $10$ unnecessary keys in the challenging ``BEG'' scenario.
Finally, note that other approaches, including \JPS, cannot be applied to find a path here as they will return no solution.

\input{fig-necessary-keys}

\vspace*{-2ex}
\paragraph{Experiment 5: Key performance overhead for unreachable destinations.}

Here, we test \invJPS over instances that are not solvable, which in fact identify the overall worst case scenario (as all the search space is explored), but also provide an upper-bound for the worst case when keys are
needed, as discussed in the previous section.
We produced $200$ paths for keys from $1$ to $10$ over the synthetic maps of Exp.~4 but we surrounded the destination by walls. The results, reported in Figure~\ref{fig:necessary-experiments}, confirm a higher runtime than all the other experiments.
For $10$ keys, the runtime is approximately $7$ times more than the ``BEG'' scenario, while it starts to show the inherent exponential nature of inventory-driven pathfinding.
Consider though that due to the symmetry-breaking benefits of \invJPS, reasoning about this scenario is nonetheless possible, while it is not feasible for \invAStar to even handle beyond $4$ keys.
Also, as happens in practice in videogames, a high runtime for unreachable paths when more keys are present can be avoided by imposing practical upper-bounds on the resources, and sacrifice completeness.

\vspace*{-2.5ex}
\paragraph{General-purpose planning}
Finally, it is possible to solve inventory-driven pathfinding by using general action-based planning \cite{GhallabNT:04-Planning}.
Such techniques in fact go beyond what \invJPS can handle, as they can mix path planning with more general dynamic reasoning, such as delivering packages or solving puzzles.
However, the performance of such general approaches is \emph{not competitive} for the particular case of inventory-driven pathfinding. We performed experiments with state-of-the-art planners \propername{FF}~\cite{Hoffmann:AIM01} and \propername{LAMA}~\cite{RichterWestphal:JAIR10-LAMA} on inventory-free maps that confirm this.
\propername{FF} could only parse maps of size up to $120\times 120$, and took a couple of seconds to synthesise paths of length below $100$.
\propername{LAMA} on the other hand was able to parse maps of size $512\times 512$ (as those used in our experiments with \invJPS), but could not solve many of those maps in several minutes.
It is important to remark that this is hardly surprising, and not even a fair comparison, since such planners rely on domain-independent heuristics and are meant to be used when known domain heuristics are not available.

%% file: fig-necessary-keys.tex
\begin{figure}[!t]
\begin{center}
\subfigure{%
\label{fig:exp5a}
\resizebox{.23\textwidth}{!}{
\begin{tikzpicture}[-]
\begin{axis}[
	-,
	xmin = 1,
	xmax = 10,
	xlabel=Number of (necessary) keys,
	ylabel=Runtime in seconds,
	scaled y ticks=false,
	ymin=0.1,
	yticklabel style={
				/pgf/number format/fixed,
				/pgf/number format/precision=2,
	},
	y label style={at={(0.02,0.5)}},
	legend columns=1,
	legend pos=north west,
	legend cell align=left
	]
	\addplot [red, mark=square]
		table [y index=1] {data/EXP2-JPS.dat};
	\addplot [blue, mark=x]
		table [y index=2] {data/EXP2-JPS.dat};
	\legend{{\invJPS, Sequential scenario}, {\invJPS, Detour scenario}}
\end{axis}
\end{tikzpicture}
}
}%
\subfigure{%
\label{fig:exp5b}
\resizebox{.23\textwidth}{!}{
\begin{tikzpicture}[-]
\begin{axis}[
	-,
	xmin = 1,
	xmax = 10,
	xlabel=Number of (necessary) keys,
	ylabel=Runtime in seconds,
	scaled y ticks=false,
	ymin=0.1,
	yticklabel style={
				/pgf/number format/fixed,
				/pgf/number format/precision=2,
	},
	y label style={at={(0.02,0.5)}},
	legend pos=north west,
	legend columns=1,
	legend pos=north west,
	legend cell align=left
	]
	\addplot [red, mark=square]
		table [y index=1] {data/FAILED.dat};
		legend pos=outer north east,
	\legend{InvJPS}
\end{axis}
\end{tikzpicture}
}
}
\end{center}
\vspace*{-3.3ex}
\caption{Performance of \invJPS when keys are necessary (left) and when no path exists (right).}
\label{fig:necessary-experiments}
\vspace*{-2ex}
\end{figure}
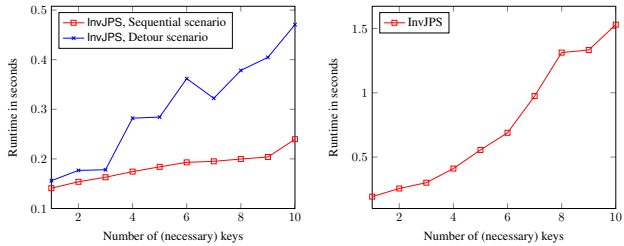

%% file: related.tex
\section{Related Work}\label{sec:related}

There are several works for improving \astar in path planning, e.g., \propername{RWA*}~\cite{Richter.etal:ICAPS10-RWA*}, \propername{RTA*}~\cite{Korf:AIJ90-RTA*}, and \propername{DAS}~\cite{Dionne.etal:SOSC11}.
While they can all easily be extended to deal with inventory-driven path planning (by just including the inventory in the state),  we expect they will all suffer the same degradation as \astar, %
as too many (``richer" in form) nodes make it to the open list, in fact exponential to the number of keys in the worst case.

As argued in~\cite{Masters:HON14}, \JPS, on the other hand, comes from a different tradition that is geared towards game programming and grid-based search with a focus on symmetry, state-space pruning, and pre-processing.
Other works in this tradition are Near-Optimal Hierarchical Pathfinding (\propername{HPA*}) \cite{BoteaMulerSchaeffer:JCD04-HPA}, \propername{Swamps} \cite{Pochter.etal:AAMAS09-SWAMPS}, and Rectangular Symmetry Reduction (\propername{RSR}) \cite{HaraborBoteaKilby:SARA11-RSR}.
In their current form, none of these existing approaches address inventory-driven path planning, but can also be
extended in a similar way.
For \propername{Swamps} it probably requires to treat inventory nodes as (intermediate) goals. For \HPA, one could consider handling the inventory only at some levels of abstraction. However, if the reasoning is done too low, completeness may be lost. %
We believe their performance will not degrade in a dramatic manner when generalizing to inventory-driven domains, however, we expect the benefits of \JPS over them---as reported in \cite{HaraborGrastien:AAAI11-JPS}---to remain unchanged against the inventory-driven versions.
As far as videogames are concerned, even though they often feature rich dynamic worlds where the
inventory-driven pathfinding scenario can easily arise, non-player characters (NPCs) are
typically unable to use their surroundings in this way. In particular about doors and passages,
what happens in practice is that either the game levels are designed in such way that this type
of interaction is irrelevant, or the NPCs have some hard-coded rules for using special objects
as dictated by game level designers (e.g., break into doors). %

%% file: conclusions.tex
\section{Conclusion and Future Work}\label{sec:conclusions}

In this paper, we introduced the problem of \emph{inventory-driven pathfinding} where the traversability of nodes is conditioned on the path taken, and generalized Jump-Point-Search (\JPS), one of the most competitive path-planning algorithms, to handle such scenarios.
The proposed algorithm, \invJPS, motivates a middle ground approach for practical deliberation mechanisms that are similar to general action-driven planning techniques, but are also \emph{grounded} to the performance achievements of sophisticated pathfinding approaches.
To the best of our knowledge there is no prior work that addresses the problem of inventory-driven pathfinding in the literature. Here we showed that the proposed algorithm \invJPS is not only a parsimonious extension of \JPS that can handle inventories, but it also preserves the performance and symmetry-breaking advantages of \JPS.

There are several extensions and optimizations we intend to explore. One of the most immediate extensions is to lift the assumption that there is no cost for picking up an item and no limit on the number of items that the agent can carry. 
As far optimizations are concerned, there are two important categories. First, optimizations that preserve optimality by re-using information from previous search that has been performed with a smaller inventory, in order to avoid some double effort. This can be done in a similar way that incremental search re-uses part of the explored search space, as e.g. in ``life-long planning \astar'' \cite{koenig2004lpa}, but in a way such that only the information that is not affected by the new item can be reused. Second, optimizations that allow computational speedups in exchange for finding possibly sub-optimal paths. For instance a variant of \invJPS may deviate from the most promising path in order to pick-up keys that are ``near'' the current location. In this way, as taking a key can only increase the reachable areas of the map, by sacrificing some optimality for picking up also keys that may not be useful at the end, we simplify the combinatorics involved and we allow for faster solutions.

%
%
%

%

%
%
%
%
%
%
%
%
%
%
%
%
%
%
%
%
%
%
%
%
%
%
%
%
%
%
%
%
%
%
%
%
%
%
%
%
%
%

%

%
%
%
%
%

%
%
%
%
%
%
%
%
%
%

%
%
%

%
%

%% file: appx-proof.tex
\appendix

\section{Proof Theorem 1}\label{appx:proof_theo1}

We first generalize Lemma 1 in \cite{HaraborGrastien:AAAI11-JPS} to account for the case of jump points due to picking keys.

\begin{lemma}
Each turning point along an optimal diagonal-first path $\pi$ is either a (standard) jump point or an inventory jump point.
\begin{proof}
Let $n_k$ be a turning point in $\pi$ and suppose that it is not in an inventory jump point.
We now consider two cases:
\begin{itemize}
     \item if the turning point is ``forward'' in the direction of travel, the three cases listed in~\cite[Lemma 1]{HaraborGrastien:AAAI11-JPS}---diagonal-to-diagonal, straight-to-diagonal, and diagonal-to-straight---to conclude that $n_k$ has to be a jump point.

     \item if, however, the turning direction is \emph{backwards} in the direction of travel, that is, back to node $n_{k+1}$ or one its two adjacent nodes also adjacent to $n_k$, then the path is \emph{not optimal} because subpath $\tuple{n_{k-1},n_k,n_{k+1}}$ in $\pi$ can be simply replaced by shorter subpath $\tuple{n_{k-1},n_{k+1}}$.
\end{itemize}
\end{proof}
\end{lemma}

Next, the proof for Theorem 1 follows almost exactly as that in~\cite{HaraborGrastien:AAAI11-JPS} but relying on the new Lemma (though we improved and elaborated it more).

\newcommand{\set}[1]{\{#1\}}

Let $\pi = \tuple{x_1,x_2,\ldots,x_n}$ be an any optimal path from start location $x_1$ to the goal location $x_n$, possibly requiring to traverse locations with keys.
As explained in~\cite{HaraborGrastien:AAAI11-JPS}, it is easy to transform that path into an equivalent length path $\pi'$ that is diagonal-first.
Next, path $\pi'$ can be split into consecutive ``stitched" subpaths $\pi'_1 \cdot \pi'_2 \cdots \pi'_k$ with each $\pi'_i = \tuple{n_i^0,\cdots,n_i^{k_i}}$, for $i \in \set{1,\ldots,k}$, representing movement on the same direction and such that $n_i^{k_i} = n_{i+1}^{0}$ (the end of a subpath is the start of the following subpath).
This means that every node $n_i^0$ at the start of each subpath is a turning point (either because it is a normal jump-point or because it is an inventory location).

By the generalized Lemma 1 above, every turning point $n_i^{k_i} = n_{i+1}^{0}$ in $\pi'$ is either a classical jump point or an inventory jump point. This means that those turning points \emph{will be inserted into the open list} in \invJPS, that is, they will \emph{not} be affected by the pruning mechanism.\footnote{Whether they are actually expanded may depend on whether this is the actual path found first during search. But what we show is that such path will be prunned and will be in the search space.}
Note that the intermediate moves $n_i^j$ to $n_i^{j+1}$ in each subpath $\pi'_i$ may not all be explicitly expanded under \invJPS procedure (they would be under an \astar search). Some of those intermediate nodes could indeed be expanded by \invJPS if they turnout out to be jump point or inventory jump point nodes but resulted in no change of direction along $\pi'$. 
In addition, as explained in~\cite{HaraborGrastien:AAAI11-JPS}, because each subpath moves along one direction, $n_i^{k_i}=n_{i+1}^0$ is always reached optimally from $n_i^0$.

Finally, the start node $n_0^0$ is always added to the open list and the goal node $n_k^{k_k}$ is by itself a standard jump point so it will also be inserted into the open list.

This has shown that if $\pi$ is an arbitrary optimal path (that \astar may find), its transformed diagonal-first path $\pi'$ will indeed be considered by \invJPS as its jump points, classical or inventory, will not be pruned away and will hence be part of the search tree.
Thus one can then apply the usual reasoning to prove optimality of \astar~\cite{HartNilssonRaphael:IEEE_TSSC68-ASTAR,RussellNorving:AIBOOK03} for tree search with admisible heuristics, and conclude that, provided there is a path to the goal only optimal paths will be returned by \invJPS.